\documentclass[acmsmall,screen]{acmart}

\AtBeginDocument{%
  \providecommand\BibTeX{{%
    \normalfont B\kern-0.5em{\scshape i\kern-0.25em b}\kern-0.8em\TeX}}}

\setcopyright{acmcopyright}
\copyrightyear{2023}
\acmYear{2023}
\acmDOI{XXXXXXX.XXXXXXX}

\usepackage{times}
\usepackage{epsfig}
\usepackage{graphicx}
\usepackage{booktabs}

\usepackage{array}
\usepackage{amsfonts}
\usepackage{amsmath}
\usepackage{multirow}
\usepackage{pifont}
\usepackage{verbatim}
\usepackage{colortbl}
\usepackage{xcolor}
\definecolor{mygray}{gray}{.9}
\usepackage{hyperref}
\begin{document}

\title{Hire: Hybrid-modal Interaction with Multiple Relational Enhancements for Image-Text Matching}

\author{Xuri Ge}

\affiliation{%
  \institution{Unviersity of Glasgow}
  \streetaddress{School of Computing Science}
  \city{Glasgow}
  \country{UK}
}
\email{x.ge.2@research.gla.ac.uk}

\author{Fuhai Chen}
\affiliation{%
  \institution{Fuzhou University}
 \streetaddress{College of Computer and Data Science}
  \city{Fu Zhou}
  \country{China}}
\email{chenfuhai3c@163.com}

\author{Songpei Xu}
\affiliation{%
  \institution{Unviersity of Glasgow}
\streetaddress{School of Computing Science}
  \city{Glasgow}
  \country{UK}
}
\email{s.xu.1@research.gla.ac.uk}

\author{Fuxiang Tao}
\affiliation{%
  \institution{ The University of Sheffield}
  \streetaddress{School of Computing Science}
  \city{Sheffield}
  \country{UK}
}
\email{f.tao.1@research.gla.ac.uk}

\author{Jie Wang}
\affiliation{%
  \institution{Unviersity of Glasgow}
\streetaddress{School of Computing Science}
  \city{Glasgow}
  \country{UK}
}
\email{j.wang.9@research.gla.ac.uk}
\author{Joemon M. Jose}
\affiliation{%
  \institution{Unviersity of Glasgow}
\streetaddress{School of Computing Science}
  \city{Glasgow}
  \country{UK}
}
\email{Joemon.Jose@glasgow.ac.uk}

\renewcommand{\shortauthors}{Ge et al.}

\begin{abstract} 
 Image-text matching (ITM) is a fundamental problem in computer vision. The key issue lies in jointly learning the visual and textual representation to estimate their similarity accurately. Most existing methods focus on feature enhancement within modality or feature interaction across modalities, which, however, neglects the contextual information of the object representation based on the inter-object relationships that match the corresponding sentences with rich contextual semantics. In this paper, we propose a Hybrid-modal Interaction with multiple Relational Enhancements  (termed \textit{Hire}) for image-text matching, which correlates the intra- and inter-modal semantics between objects and words with implicit and explicit relationship modelling. In particular, the explicit intra-modal spatial-semantic graph-based reasoning network is designed to improve the contextual representation of visual objects with salient spatial and semantic relational connectivities, guided by the explicit relationships of the objects' spatial positions and their scene graph. We use implicit relationship modelling for potential relationship interactions before explicit modelling to improve the fault tolerance of explicit relationship detection. Then the visual and textual semantic representations are refined jointly via inter-modal interactive attention and cross-modal alignment. To correlate the context of objects with the textual context, we further refine the visual semantic representation via cross-level object-sentence and word-image-based interactive attention. Extensive experiments validate that the proposed hybrid-modal interaction with implicit and explicit modelling is more beneficial for image-text matching. And the proposed \textit{Hire} obtains new state-of-the-art results on MS-COCO and Flickr30K benchmarks. 
\end{abstract}





\maketitle

\section{Introduction}
   
    Cross-modal retrieval, \textit{a.k.a.} image-text matching, aims at retrieving the most relevant images (or sentences) given a query sentence (or image), which has attracted extensive research attention in multimedia and computer vision due to its promising application, \textit{e.g.,} multimodel retrieval in searching engines, online shopping and social network.  
    Its main challenge is to encode visual and textual representations into the joint embedding space of matched images and sentences because of the heterogeneous feature representation and distribution of the two modalities.  

    To accurately measure the semantic similarity of two modalities and establish the association between two modalities, numerous methods \cite{frome2013devise, faghri2017vse++, lee2018stacked, huang2018learning, liu2019focus, wen2020learning,ge2021structured,Gradual} have been proposed to bridge the semantic gap between visual and textual representations. 
    Typically, earlier approaches \cite{frome2013devise, wang2016learning, faghri2017vse++} estimated the image-texts similarities based on the projected global visual and textual representations, which are directly extracted from the whole image and the full sentence via Convolutional Neural Networks (CNNs) and Recurrent Neural Networks (RNNs) respectively. 
    However, these rough representations are difficult to accurately identify and fully utilize high-level semantic concepts, especially those of images.
    
    Recently, many methods \cite{zhen2019deep, wen2020learning, ge2021structured,DIME, VSEwuqiong,cheng2022cross,Gradual} further take advantage of fine-grained region-level visual features from object detectors \cite{ren2015faster} with salient semantic content to enhance the high-level semantic representation of images, and align them with the word-level features of sentences. 
    These methods can be divided into two main kinds, intra-modal feature interactions \cite{wang2016learning, wen2020learning, ge2021structured,cheng2022cross,ge20243shnet} and inter-modal feature interactions \cite{lee2018stacked, huang2018learning, liu2019focus,DIME,NAAF}, to obtain a better multi-modal joint embedding space. 
   Intra-modal representation learning has been widely studied in many multi-modal tasks, such as image captioning \cite{ge2019colloquial,chen2019variational}, video caption retrieval \cite{liao2018interpretable,yang2021deconfounded}, and so on. 
    Similarly, for image-text matching, intra-modal representation learning is also important to improve the visual or textual semantic representation via the implicit and explicit semantic relationships reasoning methods within each modality, such as the graph convolution networks (GCNs) \cite{wen2020learning,GSMN,Gradual}, self-attention mechanism (SA) \cite{wu2019learning,CAMERA} and tree encoder \cite{yang2020tree,ge2021structured}, \textit{etc.} 
    For instance, \cite{wu2019learning} proposed intra-modal self-attention embeddings to enhance the representations of images or texts by self-attention mechanism, which can exploit subtle and fine-grained fragment relations in image and text, respectively. 
    \cite{li2019visual,VSRN++} proposed an implicit relationship reasoning modal based on Graph Convolutional Networks to build up connections between image regions and then generate the global visual features with semantic relationships.
    \cite{ge2021structured} developed a structured tree encoder within each modality to enhance the semantic and structural consistency representation of matched images and texts for cross-modal matching. 
    Intra-modal independent representation learning can adequately model relationships between entities within each modality via implicit or explicit reasoning approaches, which, however, fails to capture the fine-grained semantic correspondence interactions among the two modalities. 
    
    To address the above problem, many studies \cite{lee2018stacked, huang2018learning, liu2018attentive, liu2019focus,wang2019camp,NAAF} based on inter-modal interaction operations are proposed to further narrow  the semantic gaps between multiple modalities, which improve the retrieval performance by learning the accurate fine-grained visual-textual semantic correspondences between the fragments of image and text. 
    For instance, SCAN \cite{lee2018stacked} attended object regions to each word to generate the text-aware visual features for text-to-image matching and, conversely, for image-to-text matching. 
    IMRAM \cite{IMRAM} further proposed an iterative matching scheme with a cross-modal attention unit and a memory distillation unit to explore such fine-grained correspondence and refine knowledge alignments progressively. 
    \begin{figure}[t] 
    	\centering
    	\includegraphics[width=0.8\linewidth]{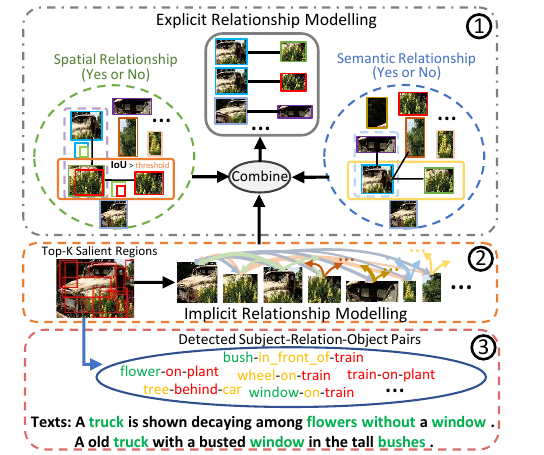}
    	\vspace{-0.5em}
    	\caption{
     \small Illustration of the explicit and implicit intra-modal modelling schemas for the semantic relationship. \ding{172} the explicit spatial-semantic relationship modelling schema: objects along with their spatial and semantic relationships are jointly modelled based on the relative position and the detected scene-graphs. 
        However, the subject-relation-object pairs (\ding{174}) in detected scene graphs of each image usually have some errors or do not match the text. For example, in \textcolor{green}{\emph{window}}-\textcolor{yellow}{\emph{on}}-\textcolor{red}{\emph{train}},  the word labels of relation ``on" and object ``train" are hard to accurately represent the corresponding semantic content, or even wrong (in red). 
        To this end, the relational connectivity (relationship exists or not) rather than the object/attribute label is encoded into the object features. 
        In addition, some relation pairs are even missing due to the limitation on the label range of the offline detector, \textit{e.g.} \emph{truck}-\emph{with}-\emph{window}. Fortunately, it can be relieved by the implicit relationship modelling (\ding{173}) due to its construction of the general relationship among object regions. 
        \ding{173} the implicit relationship modelling schema: object relationships are constructed by fully connecting the object regions, where the information can be propagated and aggregated among objects according to their potential relationships.  
        However, it is hard to maintain strong inter-object relationships in a multi-layer network. 
        To deal with the above issues, it’s intuitive to combine both implicit and explicit relationship modelling to cooperate visual semantic representation with the inter-object relationship.
        }
    	\label{fig:motivation_01}
    	\vspace{-0.5em}
    \end{figure}
    
    Moreover, recent methods \cite{GSMN,zhang2020context,wei2020multi,DIME} combined intra- and inter-modal interactions to jointly improve semantic relation representation within each modality and accurate visual-textual semantic correspondence between the two modalities, further boosting retrieval performance. 
    For instance, MMCA \cite{wei2020multi} integrated intra-modal and inter-modal interactions in a parallel pattern, in which both interactions employ implicit transformer-based self-attention mechanism \cite{vaswani2017attention}, but inter-modal interaction concatenates cross-modal region-word features for attention calculation. 
    DIME \cite{DIME} introduced a multi-layer modality interaction framework with different intra- and inter-modal interaction cells, stacked in width and depth. 
    However, the hand-crafted multi-interaction combining methods \cite{GSMN,zhang2020context,wei2020multi} lack exploration on the impact of different combinations of intra- and inter-modal interactions on matching performance and \cite{DIME}, relying on soft links and multiple interaction cell stacking, increases model complexity. 
    Additionally, these methods, despite notable improvements, overlook the limited representation of inter-object relationships compared to the strong textual context, resulting in a weakened role of visual semantics in image-text matching.
    The basic intuition of our work lies in two aspects to deal with the above problems. On the one hand, the intra-modal feature interactions, whether implicit or explicit, are crucial to enhance the visual/textual representation with the semantic relationships among fragments, especially among the visual region features that lack contextual representation. 
    However, either implicit or explicit intra-modal interactions have their own defects. 
     Notably, providing the fully-connected information flows among objects, through the implicit intra-modal interaction \cite{zhang2020context,SGRAF,cheng2022cross},  usually leaves the relationship information weak and ambiguous due to the redundant information, which affects the object discrimination as shown in Figure \ref{fig:motivation_01} (2).
    Additionally, the effect of implicit intra-modal interaction on the structured correlation among the objects and their relationships will be weakened when the object features pass the multi-layer network without further supervision. 
    Explicit intra-modal interaction heavily relies on the off-the-shelf detector \cite{yao2017boosting,wu2017image,wang2020consensus,anderson2018bottom} to concatenate the object region features with the features of the detected inter-object relationships via the graph-based modelling, which, however, introduces additional recognition error from object and attribute labels. %
    Moreover, it also neglects the spatially relative positions. 
    For instance, in \cite{wang2020cross,nguyen2021deep}, objects and their corresponding relationships are detected guided by the scene graph, and their label-based embeddings are aggregated with the object region features to feed the Graph Convolution Networks (GCNs). 
    However, due to the heterogeneous training data, the detected object and relation labels (\textit{e.g.} `\textit{train-on-plant}') are usually inconsistent with the expressions of the corresponding sentences as shown in Figure \ref{fig:motivation_01} (3). 
    To address the above issues,  it’s natural for us to integrate both the implicit and the explicit intra-modal interactions to enhance the object representation, which tackles the limitations of the structured information in implicit interactions and provides flexibility in explicit interactions. 
    To enhance object discrimination, we consider an integrated structured model to capture the explicit information of the inter-object relationships, including the semantic and spatial considerations. 
    As manifested in Figure \ref{fig:motivation_01} (1), by explicitly constructing the inter-object relationships, the semantic relationship modelling provides a strong semantic correlation between objects while the spatial relationship modelling reduces the feature redundancy of spatially overlapping.  
    Notably, we do not use the additional detected labels to mitigate the error interference from the detection and facilitate the end-to-end representation learning. 
    
    On the other hand, the effects of different combinations of the intra- and inter-modal interactions on matching results are different, which, however, are not sufficiently discussed in the existing literature \cite{GSMN,zhang2020context,wei2020multi,SGRAF,cheng2022cross,DIME,NAAF}. 
    Most of the existing hand-crafted methods combining intra-modal and inter-modal interactions directly use simple serial-pattern \cite{GSMN}, or parallel-pattern \cite{zhang2020context} combinations, which lack the discussion and exploration of different combinations. 
    Although DIME \cite{DIME} proposed a dynamic route exploration approach in multiple layers with multi-interaction, it relies on a huge serial and parallel network, which contains three layers and each layer contains four interactions. 
    In this work, we will explore, in detail, the impact of different combinations on retrieval performance, including multiple intra- and inter-modal interactions among images and sentences with explicit and implicit modelling, and discuss the potential reasons.

    Driven by the above considerations, we present a novel hybrid-modal interaction method for image-text matching via multiple relational reasoning modules within and across modalities (termed \textbf{\textit{Hire}}), which better correlates the intra- and inter-modal semantics between objects and words. 
    For the intra-modal semantic correlation, the inter-object relationships are explicitly reflected on the spatially relative positions, and the scene graph guided potential semantic relationships among the object regions. We then propose a relationship-aware GCNs model (termed \textit{R-GCNs}) to enhance the object region representations with their relationships, where the graph nodes are object region features and the graph structures are determined by the inter-object relationships, \emph{i.e.} each edge connection in the graph adjacency matrices relies on whether there is a relationship with high confidence.  
    In addition, to mitigate the impact of relation omission by the off-the-shelf detector and adequately keep structured correlations among the objects and their relationships in a multi-layer network, we perform implicit relational reasoning between objects before explicitly modelling them. 
    Experiments also prove that this information supplement effectively improves the effect of retrieval. 
    For the inter-modal semantic correlation, the implicit and explicit semantic enhanced representations of object regions, as well as the enhanced semantic representations of words that undergo a fully-connected self-attention model, are attended alternatively in the inter-modal interactive attention, where the object region features are attended to each word to refine its feature and conversely the word feature are attended to each object region to refine its feature. To correlate the context of objects with textual context, we further refine the representations of object regions and words via cross-level object-sentence and word-image-based interactive attention. 
 The intra-modal semantic correlation, inter-modal semantic correlation, and similarity-based cross-modal alignment are jointly executed to enhance the cross-modal semantic interaction further.

    The contributions of this paper are as follows: 
    \begin{itemize}
        \item  We propose an intuitive intra-model interaction model that combines implicit and explicit relationship modelling to guarantee a structured correlation among the objects and their relationships with continuous correlation guidance in a multi-layer network, overcoming the relationship omissions and erroneous via the self-attention mechanism. 
        \item  We explore an explicit intra-modal semantic enhanced correlation to utilize the inter-object spatially relative positions and inter-object semantic relationships guided by a scene graph, and propose a relationship-aware GCNs model (R-GCNs) to enhance the object region features with their relationships. This module mitigates the error interference from the detection and enables end-to-end representation learning.
        \item   We conduct exhaustive experiments on a variety of cross-modal interaction methods. Then we propose a comprehensive method (\textit{Hire}) to unite the intra-modal semantic correlation, inter-modal semantic correlation, and the similarity-based cross-modal alignment to simultaneously model the semantic correlations on three grain levels, \emph{i.e.} intra-fragment, inter-fragment, inter-instance. Especially, cross-level interactive attention is proposed to model the correlations between fragments and instances. 
        \item  The proposed \textit{Hire} is sufficiently evaluated with extensive experiments on MS-COCO and Flickr30K benchmarks and achieves a new state-of-the-art for image-text matching. 
     \end{itemize}   

    This paper is an extended version of our previous conference paper \cite{ge2023cross}, where the spatial and semantic relationship-aware GCNs are proposed to explicitly enhance object region features with the inter-object relationships, as well as cross-modal interactive refinement. 
    The main extension of this article includes three folds:
     \begin{itemize}
     \item  We combine implicit and explicit inter-object relationship modelling within visual modality, which ensures that inter-object relationships are fully explored from multiple perspectives and overcomes relationship omissions due to inaccurate offline detectors, further improving the robustness of image features.
     \item We combine the independent spatial and semantic graphs into a unified spatial-semantic graph to further mitigate the issue of partial overlapping region relationship omissions due to the detected salient object region redundancy, thereby improving the robustness of image features. 
     \item We conduct extensive experiments on MS-COCO and Flickr30K to verify the effectiveness of our proposed \textit{Hire} via a better combination of novel intra- and inter-modal interactions. We add more detailed analyses and more quantitative visualizations in terms of intra-modal relationships and cross-modal attentions, which help to interpret the behaviours of the model. In addition, we include considerable new experimental results to discuss the impact of different components and their different combinations on matching performance. 
     \end{itemize} 
     
     The remainder of this article is organized as follows. Section \ref{relatedwork} reviews the related work. Section \ref{Formulation} presents the problem formulation.  In Section \ref{Approach}, all components of our proposed \textit{Hire} are described in detail respectively. Section \ref{config} describes the datasets, evaluation metrics, and experimental configuration. In Section \ref{results}, we present the experimental results and analysis and we discuss some perspectives on large-scale trained models in Section \ref{Discussion}. Finally, we conclude the article in Section \ref{conclusion}.
     
\section{Related Work} \label{relatedwork}
    The key issue of image-text matching is to reduce the heterogeneous feature representations of the two modalities and measure the visual-text similarity between images and sentences. 
    It can be divided into three main kinds: intra-modal interactive enhanced matching, cross-modal interactive enhanced matching and hybrid-modal interactive enhanced matching. Our \textit{Hire} combine the hybrid-modal interactions for image-text matching, which includes multiple intra- and cross-modal interactive enhanced modules.
    
    \subsection{Intra-modal Interactive Enhanced Matching} 
    Most earlier works \cite{frome2013devise, kiros2014unifying, mao2014deep, fang2015captions, vendrov2015order, wang2016learning, wang2018learning} used independent intra-modal interactive processing of images and sentences within two branches to obtain a holistic representation of images and sentences. 
    Some works \cite{frome2013devise, kiros2014unifying, wang2016learning,long2024CFIR} directly extracted the features of two modalities from the whole image via CNNs and from the full sentence via RNNs to calculate the cross-modal similarities. 
    However, due to the crudeness of global features extracted from the whole images and sentences, many semantic details are ignored, especially for images with many salient object representations. 
    Inspired by the detection of object regions, many studies \cite{karpathy2014deep, wu2019learning,CAMERA} started to use the pre-extracted salient object region features to represent fine-grained images. 
    And fine-grained region-level image features and word-level text features are constructed and aligned within the modalities,  respectively. 
    For instance, DVSA in \cite{karpathy2015deep} first adopted R-CNN to detect salient objects and inferred latent alignments between word-level textual features in sentences and region-level visual features in images. 
    Furthermore, to take full advantage of high-level objects and words semantic information, many recent methods \cite{nam2017dual,wu2019learning,ge2021structured,SGRAF,VSRN++} exploited the relationships between the objects and words to help the global embedding of images and sentences, respectively. 
    For instance, 
    \cite{ge2021structured} introduced two modality parsing trees to construct structured representations of images and sentences with the explicit entity relationships in each modality tree structure. 
    Intra-modal interactive enhancement improves the cross-modal retrieval performance via relationship interactions between the objects of image and words of texts, which, however, fails to capture the fine-grained correspondence between objects and words. 
    
    \subsection{Inter-modal Interactive Enhanced Matching} 
    The fine-grained cross-modal interactive enhanced matching is widely popular to improve the visual-textual semantic alignments in many methods \cite{nam2017dual,lee2018stacked,ji2019saliency,CAMERA,zhang2020context,IMRAM,DIME}. 
   \cite{lee2018stacked} proposed a novel stacked cross-attention network to construct both image-to-text attention and text-to-image attention interactions, assigning each modality fragment with weights from another modality's fragments.  
    \cite{xu2020cross} proposed a hybrid matching method to calculate the cross-modal attention between the local fragments of two modalities for image-text matching with the help of multi-label prediction of global semantic consistency. 
    Some works \cite{wang2020cross,Gradual} employed GCNs to improve the interaction and integrate different item representations by a learned graph.   
    Gradual \cite{Gradual} introduced two-modal graphs to help the interactions between modalities, however, the post-interaction concatenation did not substantially improve interactions and additionally introduced word label noise from the scene graph. 
    And some works \cite{wang2020cross,nguyen2021deep,Gradual} also encoded the word labels from the detected visual scene graphs, causing ambiguity due to the effect of cross-domain training. 
    
    \subsection{Hybrid-modal Interactive Enhanced Matching}
    Recently, some studies \cite{wei2020multi,zhang2020context,fu2024iisan,GSMN,DIME} try to combine the intra- and cross-modal interactions to further improve the fine-grained inter-modal object-word correspondence with intra-modal interaction enhancement. 
    For instance,  
    \cite{wei2020multi} proposed a hybrid-modal relational interaction method to exploit the fine-grained relationships among the fragments via a parallel pattern of self-attention and cross-attention approaches. 
     However, the above hybrid-modal interaction methods employed implicit relationship modelling within a modality, which makes it hard to keep a structured correlation among the objects and their relationships in a multi-layer network without continuous correlation guidance. 
      The most relevant existing work to ours is DIME \cite{DIME}, which dynamically learns interaction patterns through soft-path decisions in a 4-layer network, where each layer contains two intra-modal and two inter-modal interaction strategies, respectively. 
  However, DIME relies on a large and complex network, which contains 12 units in 4 types, to assign weights to the output features of different interaction units. This makes its path selection challenging to interpret. 
    And it still suffers from the aforementioned issue of hard maintaining strong inter-object relationships in the multi-layer network. 
    
    In contrast to previous studies, \textit{e.g.}, MMCA \cite{wei2020multi}, DIME \cite{DIME}, \textit{etc.}, our \textit{Hire} approaches the inter-object modelling in a novel way by exploiting the spatial and semantic graph to enhance the structured relationship embedding based on implicit reasoning. The joint embedding space is obtained by aligning the fine-grained inter-modal semantic fragments further to reduce the heterogeneous (inter-modality) semantic gap. 
    Doing so allows us to provide more robustness than DIME \cite{DIME}, which also improves the interpretability of the model.

\section{Problem Formulation} \label{Formulation}
    Image-text matching, \textit{a.k.a.}, image-sentence retrieval, aims at matching the most relevant images in the image database (or texts in the sentence database) given a text query (or image query). 
    Here, assume we have an image database $\mathcal{I}=\{I_1,I_2,\dots,I_N\}$ and a text database $\mathcal{S}=\{S_1,S_2,\dots,S_M\}$, which contain $N$ images and $M$ sentences, respectively. 
    This paper aims to facilitate efficient image-text matching via fine-grained intra-modal relationship utilization and cross-modal semantic correspondence. 

    To this end, we first take advantage of the bottom-up-attention model \cite{anderson2018bottom} to extract top-K fine-grained sub-region features $\hat{V}= [\hat{v}_1, \dots, \hat{v}_K]$, $\hat{v}_i \in \mathbb{R}^{2048}$, for each image $I$, based on the category confidence score in an image, which can better represent the salient objects and attributes. 
    Afterwards, a fully connected (FC) layer with the parameter $W^o \in \mathbb{R}^{2048 \times D_v}$ is used to project these feature vectors into a $D_v$-dimensional space. 
    Finally, these projected object region features ${V}= [v_1, \cdots, v_K]$, $v_i \in \mathbb{R}^{D_v}$, are taken as initial visual representations without semantic relationship enhancement.
    For sentence texts, we follow the recent trends in the community of Natural Language Processing and utilize the pre-trained BERT \cite{BERT} model to extract word-level textual representations.
    Similar to visual features processing, we also utilize FC layers to project the extracted word features into a $D_t$-dimensional space for sentence $S$, denoted as $T = [t_1, t_2, \cdots, t_m]$, $t_j \in \mathbb{R}^{D_t}$, with length $m$.  
    To facilitate cross-modal interaction and embedding space consistency, we project the visual and textual representations into the same dimension ($D_v$=$D_t$). For subsequent local-global inter-modal interaction and final cross-modal similarity calculation, we use the average-pooling operation to obtain the global image feature $\bar{V}$ for text-to-image and the global text feature $\bar{T}$ for image-to-text. 
    
    Next, we leverage multiple intra-modal interactions to enhance the semantic representation within modalities and inter-modal interactions to narrow the semantic gap between heterogeneous visual-textual modalities. 
    Notably, we sufficiently explore the impact of different combinations of interactions and ultimately construct our proposed \textit{Hire}, which unite the intra-modal semantic correlation, inter-modal semantic correlation and the similarity-based cross-modal alignment together to model the semantic correlations on three levels, \textit{i.e.} intra-fragment (especially for inter-object within visual modality), inter-fragment between two modalities, and inter-instance from one modality to another modality. 
    Firstly, the visual representation ${V}$ and textual representation ${T}$ are independently enhanced by an implicit relationship interaction based on a self-attention mechanism within each modality, and an explicit spatial-semantic relationship interaction based on relationship-aware GCNs is further used to improve the visual context information among the detected salient objects in images. 
    Then, a local-local inter-modal interaction is leveraged to improve the micro consistency of the embedding space of multi-modal features via fine-grained inter-modal fragment (object-word/word-object) correlations, and a local-global inter-modal interaction is used to keep the macro consistency via similarity-based inter-instance (image-word/sentence-object) alignment. 
    Finally, the visual and textual semantic similarity is measured for the cross-modal alignment.
    
\begin{figure*}[t] 
    	\centering
    	\includegraphics[width=1\linewidth]{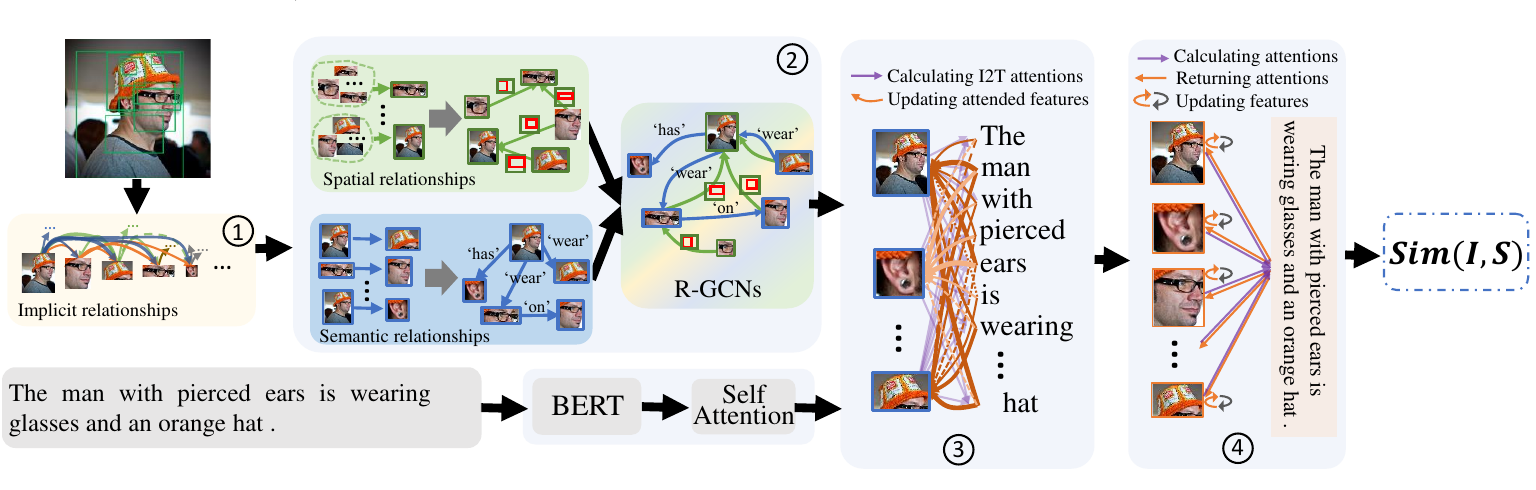}
    	\vspace{-2.7em}
    	\caption{\small The overall framework (image-to-text version) of \textit{Hire}. In intra-modal semantic correlation (\ding{172} and \ding{173}), an implicit relationship reasoning is first used to obtain the potential semantic connections among all candidate regions, similarly for high-level textual word embeddings from pre-trained BERT. And then, a relationship-aware GCNs (R-GCNs) is constructed to integrate the explicit spatial and semantic relationships between every two objects into their region representations by changing the relationship-determined graph adjacency matrix. In inter-modal semantic correlation (\ding{174} and \ding{175}), the visual and textual semantic features are further enhanced via object-word interactive attention and the visual semantic representation is refined via the cross-level object-sentence and word-image-based interactive attention. Visual and textual semantic similarity is finally estimated for the cross-modal alignment.}
    	\label{fig:fig_overall}
     \vspace{-0.6em}
    \end{figure*}      

\section{Approach} \label{Approach}
     Figure \ref{fig:fig_overall} shows the overall pipeline of our proposed \textit{Hire}, which includes two intra-modal interactions and two inter-modal interaction modules for image-text matching. For a clear presentation, we mainly describe image-to-text direction, and the text-to-image version is in a similar pattern. 
    We will first describe the intra-modal interactions for the relationship reasoning within each modality in Section \ref{intra}.  
    Afterwards, two inter-modal interaction methods are described in Section \ref{CMI} on calculating micro and macro fragment correlations from another modality. 
    Finally, the objective function is discussed in Section \ref{OBJ}.

        \subsection{Intra-modal Relationship Interactions} \label{intra} 
        Due to little inter-object relationships reflected in object representations compared to the strong context of the textual structure, we combine implicit and explicit relationship modelling approaches to improve the visual semantic representing ability. 
        The main motivation is that explicit relational graph reasoning based on the detected scene graphs maintains the inter-object relationship structure well, but suffers from relationship omission. To this end, we employ implicit inter-object relationship modelling to improve the robustness of visual representation. 
        
        \textbf{Implicit relationship modelling.}
        To refine the object-level latent embeddings of sub-region features for each image, we employ the self-attention mechanism \cite{vaswani2017attention} to concentrate on the salient information with potential correlations. 
        In particular, following \cite{vaswani2017attention}, the projected object visual features  ${V}= [v_1, \cdots, v_K]$ are used as the key and value items, and each target object $v_i$ serves as the query item. 
       Each attention weight for each query object is calculated as follows:
        \begin{equation}
             \alpha_{ij} = Att(W^q v_i, W^k v_j) = W^q v_i (W^k v_j)^T / \sqrt{D},  
        \end{equation}
        \begin{equation}
            A_{ij} = softmax(\alpha_{ij}) = \frac{exp(\alpha_{ij})}{\sum_{j=1}^K exp(\alpha_{ij})}, 
        \end{equation}
       where $W_q,W_k$ are the parameters of mapping from $D_v$ to $D$, and $\sqrt{D}$ acts as a normalization factor. Following \cite{vaswani2017attention}, we also employ multi-head dot product by $L$ parallel attention layers to speed up the calculation efficiency, and a feed-forward network (FFN) based on two FC layers (with ReLU activation function) is followed to obtain the final reasoning representation $v^A_i$ for the $i$-$th$ target object. The overall working flow is formulated as: 
        \begin{equation}
            v^A_i = FFN(W^h ||_{l=1}^L (head_1, \dots, head_L)), 
        \end{equation}
        \begin{equation}
            head_l = \sum\nolimits_{j=1}^K(A^l_{ij} W^{vl} v_j), 
        \end{equation}
        \begin{equation}
            A^l_{ij} = softmax(Att(W^{ql} v_i, W^{kl} v_j)) = softmax(W^{ql} v_i (W^{kl} v_j)^T / \sqrt{D/L}),
        \end{equation}
        where $W^h$ is the mapping parameter, $W^{ql},W^{kl},W^{vl}$ map the feature dimension to $1/L$ of the original, $||$ means concatenation. Finally, the implicit relationship enhanced visual representation $V^A = [v_1^A,\dots,v_K^A]$ is obtained.
        Similar to the above procedure, we also get the concentrated textual representation $T^A = [t_1^A,\dots,t_m^A]$ for the sentence.
        
        \textbf{Explicit visual relationship modelling.}
        To further improve the maintenance of contextual relationships among the salient objects in images, we construct a spatial-semantic graph for each image and enhance the object region features with their relationships via a relationship-aware GCNs model.  
        On the one hand, different from existing approaches \cite{li2019visual,cheng2022cross} based on implicit relationship graph reasoning, scene graphs have well-defined object relationships, which can overcome the disadvantage of fusing redundant information.
        And unlike approaches \cite{wang2020cross,Gradual} based on scene-graph enhancement, we do not encode the word labels
        predicted by the pre-trained visual scene-graph generator, like \cite{zellers2018neural}. 
        We consider that word labels from visual scene graphs of external models have errors and are semantically different from the words in the corresponding texts. 
        This tends to introduce noise that corrupts the cross-modal semantic alignment. 
        On the other hand,
        since features from the top-K candidate object regions are used for representing the image information, this leads to some regions with semantic overlap but with minor positional bias. 
        Study \cite{cheng2022cross} also indicated that the regions with larger Intersection over Union (IoU) as potentially more closely. 
        
        Different from \cite{ge2023cross,cheng2022cross}, combining spatial and semantic relationships in one graph further increases the diversity of semantic correlations, e.g. different high IoU regions with similar content can connect with some related objects which usually miss connections in the original scene graph due to confidence settings. 
        In particular, we construct an explicit spatial-semantic non-fully connected graph $\mathcal{G}=(V^A,E)$ for each image. 
        The spatial IoUs and semantic correlations between sub-regions are combined to construct the adjacency matrix $E \in \mathbb{R}^{K \times K}$ as edges for the graph. 
        Of which, if the $IoU_{ij}$ of the $i$-th region and the $j$-th region exceeds the threshold $\mu$, it indicates that there is a relationship edge between the two object regions.  Otherwise, it is $0$. 
        Likewise, if $p$-th object is associated with $j$-th object in the semantic relations extracted by a pre-trained visual scene-graph generator, there is a relationship edge between the two object regions and $0$ otherwise. 
        In this way, if the $j$-th object region has a high IoU score with $i$-th object region and semantic relationship with $p$-th object, then all three objects have associated edges with improving the robustness of relationship modelling.
        The values of edges are learning and updating based on the semantic similarities between the correlated objects, where the pairwise semantic similarity of $i$-th and $j$-th objects is calculated as: 
        \begin{equation}
            E_{ij} = (W^\varphi v_i^A)^T (W^\phi v_j^A) ,
        \end{equation}
        where $W^\varphi$ and $W^\phi$ denote the mapping parameters. 
        For simplicity, we do not explicitly represent the bias term in our paper. 
     
        For the final object region features $V^G$, the currently popular Graph Convolutional Networks (GCNS) \cite{li2019visual} with residuals are used, which can enhance the object representations by updating and embedding of spatial and semantic relationship graphs, named relationship-aware GCNs (R-GCNs), as shown in Figure \ref{fig:fig_overall}. Formally,
        \begin{equation}
        \begin{array}{c}
            V^G = (E V^A W^g) W^{r} + V^A, 
        \end{array}
        \end{equation}
        where $W^g \in \mathbb{R}^{D_v \times D_v}$ is the weight matrix of the GCN layer, $W^{r}$ is the residual weights.

    \subsection{Inter-modal Semantic Relationship Interactions} \label{CMI}    
        After image objects and text words are reinforced with semantic relationships within each modality, we apply two mainstream inter-modal interaction mechanisms to further enhance the feature representation of the target modality with attention-ware information from another modality. 
        For a clearer presentation, we describe the process as an example of image-to-text. 
        
        \textbf{Local-local inter-modal interaction.} 
        Similar to literature \cite{lee2018stacked, DIME}, we mine attention between image objects and text words to narrow the semantic gap between the two modalities. As shown in Figure \ref{fig:fig_overall} \ding{174}, taking the image-to-text example (Due to a clearer presentation), we first calculate the cosine similarities for all object-word pairs and calculate the attention weights by a  per-dimension $\lambda$-smoothed Softmax function \cite{chorowski2015attention}, as follows: 
        \begin{equation}
           c_{ij} = \frac{({v^G_i})^T t^A_j}{||v^G_i||\ ||t^A_j||}, i \in [1, K], j \in [1,m],
        \end{equation}
        \begin{equation}
           \beta_{ij} = \frac{exp(\lambda c_{ij})}{\sum_{j=1}^N exp(\lambda c_{ij})},  
        \end{equation}
        Finally, we obtain the attended object representation $v_i^F$ $\in$ $V^F$ via a conditional fusion strategy \cite{DIME} from correspondence attention-aware textual vector $q^t_i$ ($q^t_i$=$ \sum_{j=1}^m \beta_{ij} t^A$), as follows,
        \begin{equation}
            \begin{aligned}
                v_i^F = {\rm ReLU}(W^f_1(v_i^A \odot {\rm Tanh} (W^f_2 q^t_i) + W^f_3 q_i^t)) + v_i^A, \\
            \end{aligned}
        \end{equation}
        where $W^f_*$ are the mapping parameters, ${\rm ReLU}$ and ${\rm Tanh}$ are activation functions. 
        To fully explore fine-grained cross-modal interactions, we perform the above process twice. Similar, we can also obtain the word-object interaction enhancement textual features $T^F$ for the text-to-image version. 

        \textbf{Local-global inter-modal interaction.} 
        As shown in Figure \ref{fig:fig_overall} \ding{175}, we further discover the salience of the fragments in one modality guided by the global contextual information of the other modality, which makes each fragment contain more contextual features. 
        Specifically, for image-to-text, we first calculate the semantic similarity between the objects of image $V^F=\{v_1^F, \cdots, v_K^F\}$ and global textual feature $\bar{T}$. 
        Then, we can obtain the relative importance of each object via a sigmoid function. 
        Finally, we add residual connections between the attention-aware object features and the enhanced object features $V^F$, as well as the original features $V$.  
        The above process can be formulated as: 
        \begin{equation}
           r_{i} = {\rm \sigma}(W^r v_i^F \odot \bar{T}),
        \end{equation}
        \begin{equation}
          v^O_i = r_{i}v_i^F + v_i^F + {\rm ReLU} (v_i) ,
        \end{equation}
	    where $W^r$ denotes the mapping parameter. Similarly, for text-to-image, we enhance the word features by calculating the relative importance of each word between the words of the sentence and the global image feature $\bar{V}$. 
	    
	    To obtain the final match score between the image and sentence, we average and normalize the final object features of the image and calculate the cosine similarity with the global text features.

    \subsection{Objective Function} \label{OBJ}
        In the above training process, we use a bidirectional triplet ranking loss \cite{faghri2017vse++} to lead the distances between correlated image-text pairs closer than distances for uncorrelated pairs after the hybrid-modal interactions when aligning the image and sentence as follows:
        \begin{equation}
            \begin{aligned}
                \mathcal{L}_{rank}(I,{S}) = \sum_{(I,\hat{S})} [\nabla - {\rm cos}(I,S)+ {\rm cos}(I,\hat{S})]_{+} \\
                + \sum_{(\hat{I},S)} [\nabla - {\rm cos}(I,S)+ {\rm cos}(\hat{I},S)]_{+} 
            \end{aligned}
        \end{equation}     
        where $\nabla$ serves as a margin constraint, ${\rm cos}(\cdot,\cdot)$ indicates cosine similarity function, and $[\cdot]_+ = {\rm max}(0, \cdot)$. Note that $(I,S)$ denotes the given matched image-text pair, and its corresponding negative samples are denoted as $\hat{I}$ and $\hat{S}$, respectively. For image-to-text direction, $cos(I,S)= cos(V^O, \bar{T})$, and $cos(I,S)= cos(\bar{V}, {T^O})$ is for text-to-image direction.
        In addition, to preserve the semantic relevance of heterogeneous modalities in a cascaded approach consisting of multiple modules, we optimize an additional triplet ranking loss $\mathcal{L}_{add}$ for the enhanced visual and textual embeddings after the intra-modal interactions. Finally, all parameters can be simultaneously optimized by minimizing the joint bidirectional triplet ranking loss $\mathcal{L}= \mathcal{L}_{rank} + \mathcal{L}_{add}$.

\section{experimental Setup} \label{config}
    In this section, we describe our experimental setup
    , which includes the experimental datasets, the evaluation metrics, the experimental configurations and the baselines.
    \subsection{Dataset}
        We choose the most popular MS-COCO \cite{lin2014microsoft} and Flickr30k \cite{young2014f30k} datasets to evaluate our proposed model. 
        \textbf{MS-COCO}: There are over 123,000 images in MS-COCO. 
         Following the splits of most existing methods \cite{Gradual,VSRN++,VSEwuqiong,DIME}, there are 113,287 images for training, 5,000 images for validation, and 5000 for validation testing. 
        On MS-COCO, we report results on both 5-folder 1K and full 5K test sets, which are the average results of 5 folds of 1K test images and the results of full 5K test set, respectively. The full 5K test set is more challenging due to its large size.
        \textbf{Flickr30K}: There are over 31,000 images in Flickr30K with 29, 000 images for the training, 1,000 images for the testing, and 1,014 images for the validation. Since Flickr30K is smaller in diversity than MS-COCO, we initialize the network with the well-trained model from MS-COCO for further fine-tuning instead of directly training the model on Flickr30K. 
                Different AMT workers give each image in these two benchmarks five corresponding sentences. 
        
       \subsection{Evaluation Metrics}
        Quantitative performances of all methods are evaluated by employing the widely-used \cite{lee2018stacked,DIME,ge2021structured,NAAF} recall metric, R@Q (Q=1,5,10) evaluation metric, which denotes the percentage of ground-truth being matched at top Q results, respectively.
        Moreover, we report the ``\textit{rSum}'' criterion that sums up all six recall rates of R@Q, which provides a more comprehensive evaluation to testify the overall performance.

    \subsection{Implementation Details}
        Our model is trained on a single TITAN RTX GPU with 24 GB memory.
        The whole network except the Faster-RCNN model \cite{ren2015faster} is trained from scratch with the default initializer of PyTorch. The ADAM optimizer \cite{kingma2014adam} is used with a mini-batch size of 80. 
        Similar to \cite{DIME}, during the training process, we also add some negative samples from another modality for each query with the same number as the batch size.
           The learning rate is set to 0.0002 initially, with a decay rate of 0.1 every 15 epochs. The maximum epoch number is set to 30. 
        The margin of triplet ranking loss $\nabla$ is set to 0.2. 
        The threshold $\mu$ is set to 0.4. 
        For the visual object features, Top-K (K=36) object regions are selected with the highest class detection confidence scores. 
        The visual scene graphs are generated by Neural Motifs \cite{zellers2018neural}, and we use the maximum IoU to find the corresponding regions in the original Top-K salient regions. 
       The textual features are extracted by a basic version of the pre-trained 12-layer BERT with a hidden size of 768.
           The initial dimensions of visual and textual embedding space are set to 2048 and 768, respectively, which are transformed to the same 1024-dimensional (\textit{i.e.}, $D_v$= $D_s$=1024).  
      Most dimensions of mapping parameters are set to 256-dimensional (D=256) for the joint embedding space. 
        We use 16 (L=16) parallel attention layers in multi-head operations. 
           Similar to \cite{DIME}, the $\lambda$ is set to 4 in the image-to-text direction and nine in the text-to-image direction. 
        During the training process, we randomly mask 10\% words of each sentence. 
\begin{table*}[t]
\scriptsize
\begin{center}
\fontsize{9.5}{13}\selectfont
\renewcommand\tabcolsep{4.3pt}
\caption{Comparisons of experimental results on MS-COCO 5-folds 1K test set. $^*$ indicates the performance of an ensemble model. $^\dagger$ denotes the significant improvements on R@1 (paired t-test, p < 0.01) compared with the best baseline (\textit{i.e.} AME$\rm ^*$). Red numbers denote the improvements compared with state-of-the-arts.} \label{tab:tab1_coco_1k}  
\begin{tabular}{c|cccccc|c}
\hline

\hline
\multicolumn{1}{c|}{\multirow{2}{*}{Method}} & \multicolumn{3}{c}{Image-to-Text} & \multicolumn{3}{c|}{Text-to-Image}      & \multirow{2}{*}{rSum} \\
\multicolumn{1}{c|}{}                        & R@1        & R@5       & R@10      & R@1  & R@5 & \multicolumn{1}{c|}{R@10} &                        \\ \hline


    IMRAM$\rm ^*_{CVPR'20}$ \cite{IMRAM}      & 76.7    & 95.6    & 98.5    & 61.7    &   89.1   & 95.0  & 516.6     \\ 
    CAAN$\rm _{CVPR'20}$ \cite{zhang2020context}       & 75.5    &   95.4     &  98.5    &  61.3    &   89.7    & {95.2}  & 515.6     \\ 
    GSMN$\rm ^*_{CVPR'20}$ \cite{GSMN} & 78.4    & 96.4   & 98.6     & 63.3     & 90.1     &  95.7      & 522.5  \\
    
    SMFEA$\rm _{ACMMM'21}$ \cite{ge2021structured}  & 75.1    & 95.4    & 98.3     & 62.5     & 90.1     & 96.2     & 517.6  \\
    SGRAF$\rm ^*_{AAAI'21}$ \cite{SGRAF}  & 79.6    & 96.2    & 98.5     & 63.2     & 90.7     & 96.1     & 524.3  \\
    VSE$\rm \infty_{CVPR'21}$ \cite{VSEwuqiong} & 79.7    & 96.4    & 98.9     & 64.8     & 91.4     & 96.3     & 527.5  \\
    DIME$\rm ^*_{SIGIR'21}$ \cite{DIME}   & 78.8     & 96.3    & 98.7     & 64.8     & 91.5     & 96.5     & 526.6 \\ 
    VSRN++$\rm ^*_{TPAMI'22}$ \cite{VSRN++}  & 77.9    & 96.0     & 98.5   &  64.1     & 91.0   & 96.1  & 523.6  \\ 
    GraDual$\rm ^*_{WACV'22}$ \cite{Gradual} & 77.0    & 96.4   & 98.6     & {65.3}     & {91.9}     & 96.4      & 525.6  \\ 
    NAAF$\rm ^*_{CVPR'22}$ \cite{NAAF} & {80.5}    & 96.5     & 98.8     & 64.1     & 90.7     & {96.5}      & 527.2  \\ 
    AME$\rm ^*_{AAAI'22}$ \cite{li2022action}    & 79.4   & \textbf{96.7}    & {98.9}     & {65.4}     & 91.2     & 96.1      & 527.7  \\ 
    RCTRN$\rm ^*_{ACMMM'23}$ \citep{li2023reservoir}   & 79.4 & 96.6 & 98.3  &  \textbf{66.9} & \underline{92.2} &  \underline{96.8} &  \underline{530.2} \\
    KIDRR$\rm ^*_{IP\&M'23}$ \citep{xie2023unifying} &  \underline{80.9} & 96.5 &  \textbf{99.0} & 65.0 & 91.1 & 96.1 & 528.6 \\

  \hline
  
\color{gray}{CMSEI$^*$ }  & \color{gray}{81.4}    & \color{gray}{96.6}      & \color{gray}{98.8}     & \color{gray}{65.8}     & \color{gray}{91.8}     & \color{gray}{96.8}      & \color{gray}{531.1}  \\  
\multicolumn{1}{c|}{\textit{Hire}$^*$ (ours)}  & \textbf{81.6$^\dagger$\textcolor{red}{$_{+0.7}$}}    & \underline{96.6\textcolor{red}{$_{-0.1}$}}      & \textbf{99.0\textcolor{red}{$_{+0.0}$}}     & \underline{66.4\textcolor{red}{$_{-0.5}$}}     & \textbf{92.3\textcolor{red}{$_{+0.1}$}}     & \textbf{96.8\textcolor{red}{$_{+0.0}$}}      & \textbf{532.6$^\dagger$\textcolor{red}{$_{+2.4}$}}  \\  
\hline

\hline
\end{tabular}
\end{center}
\end{table*}

    \subsection{Comparison with State-of-the-art Methods}
        We compare our proposed \textit{Hire} with three kinds of image-text matching methods, including (1) intra-modal interaction-based, inter-modal interaction-based and hybrid-modal interaction-based methods. 
        \begin{itemize}
        \item Intra-modal interaction-based methods:  SGRAF \cite{SGRAF}, VSRN \cite{li2019visual}, VSE$\infty$ \cite{VSEwuqiong} (the reported version with same object inputs), SMFEA\cite{ge2021structured}, VSRN++ \cite{VSRN++}, AME \cite{li2022action}, and CHAN \citep{pan2023fine} \textit{etc}. These methods focus on  feature enhancement via relationship reasoning within an independent modality.
        \item Inter-modal interaction-based methods: SGRAF \cite{SGRAF}, CAAN \cite{zhang2020context}, IMRAM \cite{IMRAM}, NAAF \cite{NAAF}, and  RCTRN*\citep{li2023reservoir}. These methods focus on the multi-modal attention mechanism to explore the cross-modal fine-grained semantic correspondences. 
        \item Hybrid-modal interaction-based methods: CAAN \cite{zhang2020context}, GraDual \cite{Gradual}, and DIME \cite{DIME}. These methods combine intra- and inter-modal interactions to enhance the visual and textual representations via intra-modal relationship modelling and inter-modal fragment attention modelling.
        \end{itemize}

\begin{table*}[t]
\scriptsize
\begin{center}
\fontsize{9}{12}\selectfont
\renewcommand\tabcolsep{4.5pt}
\caption{Comparisons of experimental results on MS-COCO 5K test set. $^*$ indicates the performance of an ensemble model. $^\dagger$ denotes the statistical significance for p < 0.01 over R@1 compared with the best baseline (\textit{i.e.} AME$\rm ^*$). Red numbers denote the improvements compared with state-of-the-arts.} \label{tab:tab1_coco_5k} 
\begin{tabular}{c|cccccc|c}
\hline

\hline
\multicolumn{1}{c|}{\multirow{2}{*}{Method}} & \multicolumn{3}{c}{Image-to-Text} & \multicolumn{3}{c|}{Text-to-Image}      & \multirow{2}{*}{rSum} \\
\multicolumn{1}{c|}{}                        & R@1        & R@5       & R@10      & R@1  & R@5 & \multicolumn{1}{c|}{R@10} &                        \\ \hline
         
    VSRN$\rm ^*_{ICCV'19}$ \cite{li2019visual}    & 53.0  & 81.1   & 89.4  & 40.5    & 70.6   & 81.1     & 415.7  \\ 
    IMRAM$\rm ^*_{CVPR'20}$ \cite{IMRAM}   & 53.7 &  83.2  &  91.0 & 39.7    & 69.1   & 79.8     & 416.5    \\ 
    CAAN$\rm _{CVPR'020}$ \cite{zhang2020context}   & 52.5 &  83.3    & 90.9   & 41.2  &  70.3   & 82.9  &  421.1   \\ 
    VSE$\rm \infty_{CVPR'21}$ \cite{VSEwuqiong}  & 58.3   & 85.3 & {92.3}    & 42.4     &  72.7   &  \underline{83.2}     & 434.3  \\
    DIME$\rm _{SIGIR'21}$ \cite{DIME}    & {59.3}   & {85.4}   & 91.9    & {43.1}   &  \underline{73.0}   &  83.1     & {435.8}\\ 
    VSRN++$\rm ^*_{TPAMI'22}$ \cite{VSRN++}   & 54.7 & 82.9   & 90.9    & 42.0   & 72.2    & 82.7     & 425.4  \\ 
    NAAF$\rm ^*_{CVPR'22}$ \cite{NAAF}    & 58.9   & 85.2   & 92.0   & 42.5   & 70.9   & 81.4   &  430.9  \\ 
    AME$\rm ^*_{AAAI'22}$ \cite{li2022action}    &59.9   & {85.2}    & {92.3}     & \underline{43.6}     & 72.6     & 82.7      & {436.3}  \\ 
    
 RCTRN$\rm ^*_{ACMMM'23}$ \citep{li2023reservoir} &  57.1 &  83.4 &  91.9 &  43.6 &  71.9 &  83.7  &  431.6 \\
 KIDRR$\rm ^*_{IP\&M'23}$ \citep{xie2023unifying}  &  \underline{60.3} &  \underline{86.1} &  \underline{92.5} & 43.5 & 72.8 & 82.8 &  \underline{438.0} \\
\hline
\color{gray}{CMSEI$^*$}  & \color{gray}{61.5}    & \color{gray}{86.3}      & \color{gray}{92.7}     & \color{gray}{44.0}     & \color{gray}{73.4}     & \color{gray}{83.4}      & \color{gray}{441.2}  \\
\multicolumn{1}{c|}{\textbf{\textit{Hire}$^*$ (ours)}}  & \textbf{61.7$^\dagger$\textcolor{red}{$_{+1.4}$}}    & \textbf{86.7\textcolor{red}{$_{+0.6}$}}      & \textbf{92.8\textcolor{red}{$_{+0.3}$}}     & \textbf{45.2$^\dagger$\textcolor{red}{$_{+1.6}$}}     & \textbf{74.5\textcolor{red}{$_{+1.5}$}}     & \textbf{84.2\textcolor{red}{$_{+1.0}$}}      & \textbf{445.0$^\dagger$\textcolor{red}{$_{+7.0}$}}  \\ 
\hline

\hline
\end{tabular}
\end{center}
\end{table*}
\begin{table*}[t]
\scriptsize
\begin{center}
\fontsize{9}{12}\selectfont
\renewcommand\tabcolsep{4.3pt}
\caption{Comparisons of experimental results on Flickr30K 1K test set. '$^*$' indicates the performance of an ensemble model. $^\dagger$ denotes the statistical significance for p < 0.01 over R@1 compared with the best baseline (\textit{i.e.} AME$\rm ^*$)} \label{tab:tab1_F30k}
\begin{tabular}{c|cccccc|c}
\hline

\hline
\multicolumn{1}{c|}{\multirow{2}{*}{Method}} & \multicolumn{3}{c}{Image-to-Text} & \multicolumn{3}{c|}{Text-to-Image}      & \multirow{2}{*}{rSum} \\
\multicolumn{1}{c|}{}                        & R@1        & R@5       & R@10      & R@1  & R@5 & \multicolumn{1}{c|}{R@10} &                        \\ \hline 
    CAAN$\rm _{CVPR'20}$ \cite{zhang2020context}       & 70.1          & 91.6          & 97.2  & 52.8          & 79.0          & 87.9          & 478.6                 \\ 
    GSMN$\rm ^*_{CVPR'20}$ \cite{GSMN} & 76.4    & 94.3   & 97.3     & 57.4     & 82.3     &  89.0      & 496.8  \\
    SMFEA$\rm _{ACMMM'21}$ \cite{ge2021structured}  & 73.7    & 92.5    & 96.1     & 54.7     & 82.1     & 88.4     & 487.5  \\
    SGRAF$\rm ^*_{AAAI'21}$ \cite{SGRAF} & 77.8    & 94.1   & 97.4     & 58.5     & 83.0     & 88.8      & 499.6  \\
    DIME$\rm ^*_{SIGIR'21}$ \cite{DIME} & 81.0    & \underline{95.9}   & {98.4}     & 63.6     & {88.1}     & {93.0}      &  {520.0}  \\ 
    
    VSRN++$\rm ^*_{TPAMI'22}$ \cite{VSRN++} & 79.2    & 94.6   & 97.5     & 60.6     & 85.6     & 91.4      & 508.9  \\ 
    GraDual$\rm ^*_{WACV'22}$ \cite{Gradual} & 78.3    & 96.0   & 98.0     & {64.0}     & 86.7     & 92.0      & 511.4  \\ 
    NAAF$\rm ^*_{CVPR'22}$ \cite{NAAF} & {81.9}    & 96.1   & 98.3     & 61.0     & 85.3     & 90.6      & 513.2  \\ 
    AME$\rm ^*_{AAAI'22}$ \cite{li2022action}    & \underline{81.9}   & {95.9}    & \underline{98.5}     & \underline{64.6}     & \underline{88.7}     & \underline{93.2}      & \underline{522.8}  \\ 
 CHAN$\rm _{CVPR'23}$ \citep{pan2023fine}   &  80.6 &  96.1 &  97.8 &  63.9 &  87.5 &  92.6 &  518.5 \\
 RCTRN$\rm ^*_{ACMMM'23}$ \citep{li2023reservoir}  &  78.4  &  95.4 &  96.8 &  60.4 &  84.9 &  93.7 &  509.6 \\
 KIDRR$\rm ^*_{IP\&M'23}$ \citep{xie2023unifying}  & 80.2 & 94.9 & 98.0 & 61.5 & 84.5 & 90.1 & 509.2 \\
    \hline
   \color{gray}CMSEI$^*$      & \color{gray}{82.3}  & \color{gray}{96.4} & \color{gray}{98.6}  & \color{gray}{64.1} & \color{gray}{87.3} & \color{gray}{92.6} & \color{gray}{521.3}  \\
    \textbf{\textit{Hire}$^*$ (ours) }   & \textbf{83.0$^\dagger$\textcolor{red}{$_{+1.1}$}}  & \textbf{97.0\textcolor{red}{$_{+1.1}$}} & \textbf{98.8\textcolor{red}{$_{+0.3}$}}  & \textbf{65.9$^\dagger$\textcolor{red}{$_{+1.3}$}} & \textbf{89.1\textcolor{red}{$_{+0.4}$}} & \textbf{93.4\textcolor{red}{$_{+0.2}$}} & \textbf{527.1$^\dagger$\textcolor{red}{$_{+4.3}$}}\\ \hline  
    
\hline
\end{tabular}
\end{center}
\end{table*}

\section{EXPERIMENTAL RESULTS AND DISCUSSION} \label{results}
    In this section, we report the results of our experiments to evaluate the proposed approach, \textit{Hire}. 
    Note that some ensemble models with ``*" are further improved due to the complementarity between multiple models. For a fair comparison, we also provide the ensemble results in Table \ref{tab:tab1_coco_1k}, Table \ref{tab:tab1_coco_5k}, and Table \ref{tab:tab1_F30k}, which are averaged similarity scores of image-to-text version and text-to-image version. 

        \subsection{Quantitative Comparison on MS-COCO.}
        \noindent \textbf{On 5-folds 1K dataset.}
        Table \ref{tab:tab1_coco_1k} presents the experimental results compared with the previous methods on MS-COCO 5-folds 1K. 
        Specifically, compared with the best intra-modal interaction-based method KIDRR* \cite{xie2023unifying}, our \textit{Hire} obtains a significant improvement on most metrics, \textit{e.g.,} 81.6\% \textit{vs.} 80.9\% and 66.4\% \textit{vs.} 65.0\% on R@1 for image-to-text and text-to-image, respectively.
        Compared with the best inter-model interaction model RCTRN* \citep{li2023reservoir} on MS-COCO 1K test set, our \textit{Hire} achieves 2.4\% improvements in terms of rSum. 
        Compared with the best hybrid-modal interaction method DIME \cite{DIME}, which also combines multiple intra- and inter-model interactions in a multi-layer network, our \textit{Hire} achieves higher results on all metrics, \textit{e.g.,} 81.6\% \textit{vs.} 78.8\% and 66.4\% \textit{vs.} 64.8\% in terms of R@1 for text retrieval and image retrieval, respectively. 
        And \textit{Hire} clearly outperforms the methods GraDual \cite{Gradual} and KIDRR* \cite{xie2023unifying}, which also employ graph networks, by 7.0\% and 4.0\% in terms of \textit{rSum}, respectively.
        
        \noindent \textbf{On Full 5K dataset.}
        On the larger image-text matching test data (MS-COCO Full 5K test set), including 5000 images and 25000 sentences, \textit{Hire} obtains a significant improvement on all metrics compared with recent methods. 
        Compared with the latest state-of-the-arts AME \cite{li2022action}, RCTRN* \citep{li2023reservoir} and  KIDRR* \citep{xie2023unifying} , our \textit{Hire} achieves 8.7\%, 13.4\% and 7\% improvements in terms of \textit{rSum} via the common protocol \cite{VSRN++,NAAF,li2022action}, respectively. 
        And compared with the best hybrid-modal interaction method DIME \cite{DIME}, \textit{Hire} also demonstrates superiority (\textit{e.g.,} 61.7\% \textit{vs.} 59.3\% on R@1 of text retrieval and 45.2\% \textit{vs.} 43.1\% on R@1 of image retrieval).
        It clearly demonstrates the powerful effectiveness of the proposed \textit{Hire} model with the huge improvements.  

         \subsection{Quantitative Comparison on Flickr30K}
        The experimental results on the Flickr30k dataset are shown in Table \ref{tab:tab1_F30k}. From Table \ref{tab:tab1_F30k}, we can observe that our \textit{Hire} outperforms all its competitors with impressive margins on all metrics. 
        In particular, compared with the state-of-the-art method AME \cite{li2022action}, \textit{Hire} achieves higher results on all metrics (over 1.1\% and 1.3\% on R@1 for text retrieval and image retrieval, and higher 4.3\% in terms of \textit{rSum}). 
        In addition, compared with the most relevant existing work DIME \cite{DIME}, \textit{Hire} achieves 2.0\%, 2.3\% and 7.1\% improvements of R@1 on image-to-text, R@1 on text-to-image and \textit{rSum}, respectively.

        \subsection{Generalization Ability for Domain Adaptation} 
     We further validate the generalization ability of the proposed \textit{Hire} on challenging cross-datasets (It means training the model on one dataset and testing the model on another), which is meaningful for evaluating the cross-modal retrieval performance in real-scenario. 
        Specifically, similar to CVSE \cite{wang2020consensus}, we transfer our model trained on MS-COCO to Flickr30K dataset. 
        As shown in Table \ref{tab:tab_transfor}, the proposed \textit{Hire} has an impressive advantage in cross-modal retrieval compared with its competitors. 
        For instance, compared with the best method DIME \cite{DIME}, \textit{Hire} achieves significantly outperforms on R@1 of text retrieval, R@1 of image retrieval, and \textit{rSum} with 4.2\%, 1.3\% and 8.4\% improvements, respectively. 
               It reflects that \textit{Hire} has excellent generalisation capability for cross-dataset image-text matching. 

\begin{table}[t]
\scriptsize
\begin{center}
\fontsize{9}{11.5}\selectfont
\renewcommand\tabcolsep{6.0pt}
\caption{Comparison results on cross-dataset generalization from MS-COCO to Flickr30k.  $^{\natural}$ means the results are obtained from their published pre-trained model.  $^\dagger$ denotes the statistical significance for p < 0.01 over R@1 compared with the best baseline (\textit{i.e.} DIME$\rm ^*$)} \label{tab:tab_transfor}
\begin{tabular}{c|cccccc|c}
\hline

\hline
\multicolumn{1}{c|}{\multirow{2}{*}{Method}} & \multicolumn{3}{c}{Image-to-Text} & \multicolumn{3}{c|}{Text-to-Image}      & \multirow{2}{*}{rSum} \\
\multicolumn{1}{c|}{}                        & R@1        & R@5       & R@10      & R@1  & R@5 & \multicolumn{1}{c|}{R@10} &                        \\ \hline

    VSE++$\rm _{BMVC'18}$ \cite{faghri2017vse++}       & 40.5 & 67.3 & 77.7	& 28.4	& 55.4 & 66.6 & 335.9\\
    LVSE$\rm _{CVPR'18}$ \cite{engilberge2018finding}   & 46.5 & 72.0 & 82.2	& 34.9	&62.4  & 73.5 & 371.5\\
    SCAN$\rm ^*_{ECCV'18}$ \cite{lee2018stacked}	   & 49.8 & 77.8 & 86.0	& 38.4	& 65.0 & 74.4 & 391.4\\
    CVSE$\rm _{ECCV'20}$ \cite{wang2020consensus}	   & 56.4 & 83.0 & 89.0	& 39.9	& 68.6 & 77.2 & 414.1\\
    VSE$\rm \infty^{\natural}_{CVPR'21}$ \cite{VSEwuqiong}  & \underline{68.0} &89.2	& 93.7	& 50.0	& 77.0 & 84.9 & 462.8 \\
    DIME$\rm ^{*\natural}_{SIGIR'21}$ \cite{DIME}  & 67.4	& \underline{90.1} & \underline{94.5}	& \underline{53.7}	& \underline{79.2} & \underline{86.5}  & \underline{471.4}\\
    \hline 	
    \color{gray}CMSEI$^*$      & \color{gray}{69.6}  & \color{gray}{89.2} & \color{gray}{95.2}  & \color{gray}{53.7} & \color{gray}{79.5} & \color{gray}{87.2} & \color{gray}{474.4}  \\
    \textbf{\textit{Hire}$^*$ (ours) }      & \textbf{71.6$^\dagger$\textcolor{red}{$_{+3.6}$}}  & \textbf{90.5\textcolor{red}{$_{+0.4}$}} & \textbf{95.2\textcolor{red}{$_{+0.7}$}}  & \textbf{55.0$^\dagger$\textcolor{red}{$_{+1.3}$}} & \textbf{80.1\textcolor{red}{$_{+0.9}$}} & \textbf{87.4\textcolor{red}{$_{+0.9}$}} & \textbf{479.8$^\dagger$\textcolor{red}{$_{+8.4}$}} \\ 
    \hline 
    
\hline
\end{tabular}
\end{center}
\end{table}

    \subsection{Ablation Studies}
    In this subsection, we perform detailed ablation studies in Table \ref{tab:tab3_ab} on the MS-COCO 5-folds 1K test set to evaluate the effectiveness of each component in our proposed \textit{Hire}. And we also explore and discuss the impact of different combinations of multiple intra- and inter-modal interactions on the effectiveness of cross-modal retrieval.  
    
    \textbf{Effects of visual-textual implicit reasoning.} 
      In Table \ref{tab:tab3_ab}, the performance of \textit{Hire} drops from  532.6\% to 529.1\% and to 531.4\%, when removing the visual and textual implicit reasoning model (indicated by w/o VSA or w/o TSA), respectively. 
      When removing the self-attention-based implicit reasoning model, it degrades the R@1 score by 0.5\% and 0.4\% on image-to-text and text-to-image, and reduces 1.4 \% in terms of rSum.  
      These observations suggest that implicit attention can slightly improve the information concentration between the fragments within each modality. 
      
    \textbf{Effects of visual spatial-semantic graph reasoning.} 
    In Table \ref{tab:tab3_ab}, \textit{Hire} decreases absolutely by 6.2\% on MS-COCO 5-fold 1K test set in terms of $rSum$ when removing the visual spatial-semantic graph (w/o VSSG). 
    It suggests that spatial-semantic graph reasoning plays an important role in concentrating on relevant regional fragment features, both spatially and semantically. 
    In addition, compared with CMSEI \cite{ge2023cross}, which split the spatial and semantic relationships into two separate graphs, our \textit{Hire} increases 4.5\% in terms of $rSum$ on MS-COCO. 
    It demonstrates that the integration of spatial and semantic relationships can further improve the effective construction of fragment relationships and improve the robustness of the model. 

    \textbf{Effects of explicit textual graph reasoning.} 
    We also model explicit relationships existing in the text to explore their effects. Specifically, we apply the Stanford enhanced dependency parser \cite{chen2014fast} following  \cite{Gradual} to extract the explicit textual scene graph and use the same R-GCN module as the vision component to model its relationship. However, when adding the textual R-GCN into our model, the matching performance drops from 532.6 to 529.5 in terms of rSum. We speculate that the main reason is that the original sentence already provides richer contextual information than the parsed textual scene graph, where the parsed textual scene graph is incomplete due to the lack of some attributes during the parsing process.
    
\begin{table}[t]
\scriptsize
\begin{center}
\fontsize{9}{12}\selectfont
\renewcommand\tabcolsep{10.0pt}
\caption{Ablation studies on MS-COCO 1K test set. All values are ensemble results by averaging two models' (I-T and T-I) similarity. CMSEI$^*$(w/o) means that the spatial-semantic graph is split into two separate graphs, as well as lacking textual semantic enhancement.} \label{tab:tab3_ab}
\begin{tabular}{c|cccccc|c}
\hline

\hline
\multicolumn{1}{c|}{\multirow{2}{*}{Method}} & \multicolumn{3}{c}{Image-to-Text} & \multicolumn{3}{c|}{Text-to-Image}  & \multirow{2}{*}{rSum} \\
\multicolumn{1}{c|}{}                        & R@1        & R@5       & R@10      & R@1  & R@5 & \multicolumn{1}{c|}{R@10}  \\ \hline
    \textit{Hire}       & \textbf{81.6}    & \textbf{96.6}   & \textbf{98.9}     & \textbf{66.4}     & \textbf{92.3}     & \textbf{96.8}      & \textbf{532.6}   \\ \hline
    w/o VSA         &  81.3  &  96.2    & 98.4    & 65.3    & 91.6    & 96.3  & 529.1 \\
    w/o TSA         &  81.5   &  96.3    &  98.6   & 66.2    & 92.3     &  96.5  &531.4  \\
    w/o SA           &  81.1 &  96.5    & 98.7    & 66.0    & 92.2     & 96.7   & 531.2  \\
    CMSEI$^*$(w/o) \cite{ge2023cross}  &  80.9  &  96.0    & 98.2    & 65.1    & 91.5     & 96.4   & 528.1  \\
    w/o VSSG          &  80.1   &  96.2    & 98.1    & 64.1    & 91.5    & 96.4    & 526.4     \\
    w/o LLII         &  79.2   &  95.7    & 97.6    &  64.2   &  91.0    &  95.5    & 523.2       \\
    w/o LGII         &  81.1   &  96.6    &  98.8  & 66.0     &  92.2    &  96.5  & 531.2   \\
    \hline
    
\hline
\end{tabular}
\end{center}
\end{table}  

\begin{table}[t]
\scriptsize
\vspace{-0.5em}
\begin{center}
\fontsize{9}{12}\selectfont
\renewcommand\tabcolsep{10.0pt}
\caption{Performance comparison of component orders on MS-COCO 1K test set. All values are ensemble results by averaging two models' (I-T and T-I) similarity.} \label{tab:tab3_comb}
\begin{tabular}{c|cccccc|c}
\hline

\hline
\multicolumn{1}{c|}{\multirow{2}{*}{Combination}} & \multicolumn{3}{c}{Image-to-Text} & \multicolumn{3}{c|}{Text-to-Image}  & \multirow{2}{*}{rSum} \\
\multicolumn{1}{c|}{}                        & R@1        & R@5       & R@10      & R@1  & R@5 & \multicolumn{1}{c|}{R@10}  \\ \hline
    \begin{tabular}[c]{@{}c@{}}\textit{Hire}\\ $\mathcal{A}$(\ding{172}\ding{173}) $\mathcal{B}$(\ding{174}\ding{175})\end{tabular}      & \textbf{81.6}    & \textbf{96.6}   & \textbf{98.9}     & \textbf{66.4}     & \textbf{92.3}     & \textbf{96.8}      & \textbf{532.6}   \\ \hline
    $\mathcal{B}$(\ding{174}\ding{175}) $\mathcal{A}$(\ding{172}\ding{173})      &  71.4  &  90.8 &  92.7  &  64.4  &  91.1  &  96.3  &  506.7 \\
    $\mathcal{A}$(\ding{173}\ding{172}) $\mathcal{B}$(\ding{174}\ding{175})          &  81.1   &  96.0    &  98.7   & 66.0    & 91.8     &  96.2  & 529.8\\
    $\mathcal{A}$(\ding{172}\ding{173}) $\mathcal{B}$(\ding{175}\ding{174})  &  81.4  &  96.6 &  98.8  &  66.1  &  92.2  &  96.7  &  531.8 \\
    \hline
    
\hline
\end{tabular}
\end{center}
\end{table}

\begin{figure}[t] 
	\centering
 \includegraphics[width=0.9\linewidth]{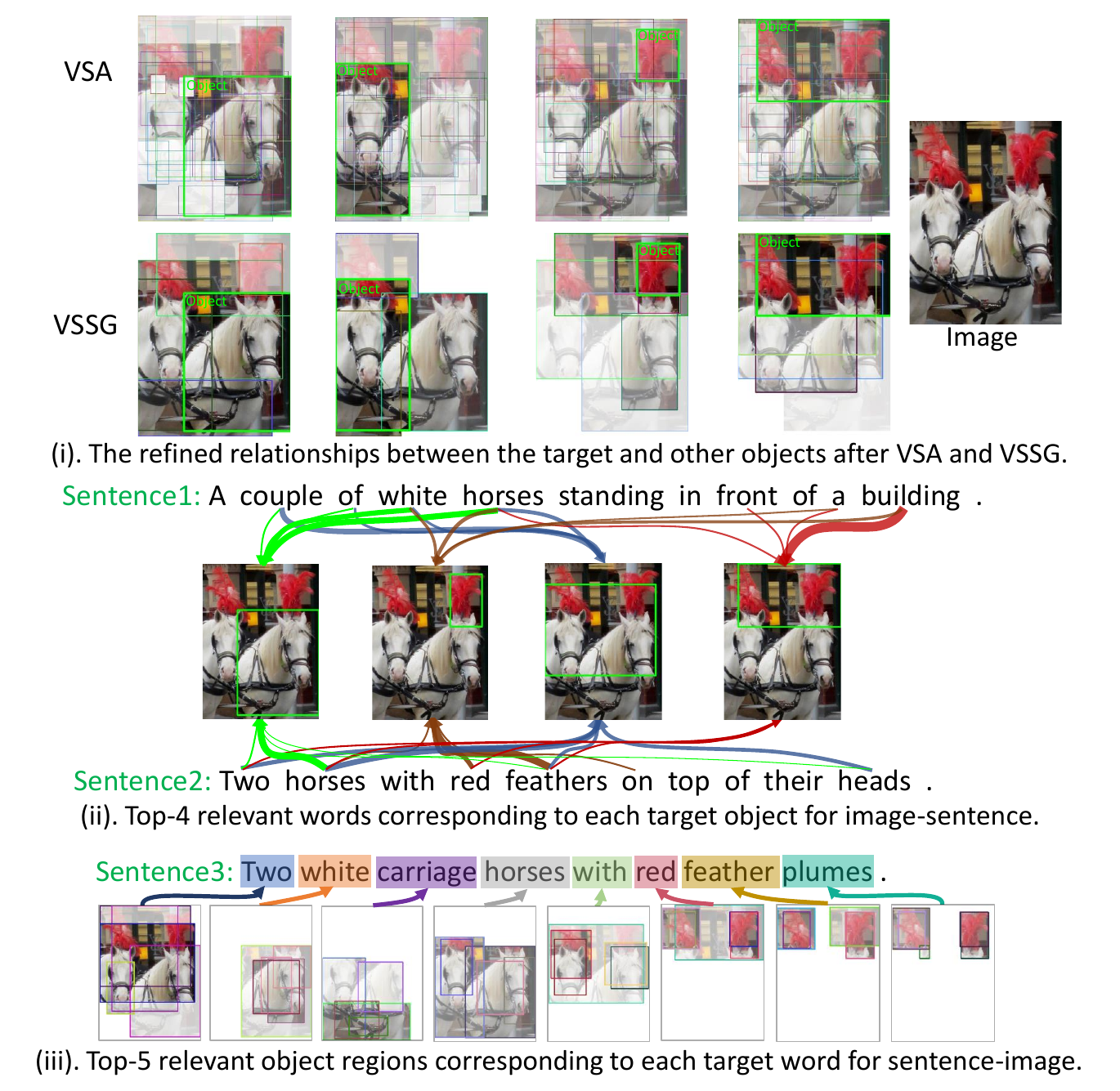}
	\vspace{-1em}
	\caption{Visualization of main modules: (i) the refined relationships between the target object (in green box) and other correlated object regions after implicit visual relationship reasoning (VSA) and explicit visual spatial-semantic graph reasoning (VSSG), (ii) results on top-4 region-words pair correspondences of each target object (in green box) for image-to-text, (iii) results on top-5 word-regions pair correspondences of each target word for text-to-image. The degree of white coverage of regions and the thickness of lines indicate different learning weights (best viewed in color).} 
	\label{fig:example_Graph_I2TLL_T2ILL_attention}
\end{figure}

    \textbf{Effects of local-local and local-global inter-modal interactions.} 
    We evaluate the impact of the local-local and local-global inter-modal interaction (LLII and LGII) for \textit{Hire}. 
    As shown in Table \ref{tab:tab3_ab}, the absence of LLII and the absence of LGII reduce 9.4\% and 1.4\% in terms of rSum on MS-COCO 5-folds 1K test set, respectively. 
    It is obvious that the multiple inter-modal interactions play a vital role in image-text matching process, which also suggests that cross-modal interactions effectively narrow the semantic gap between the two modalities.

    \textbf{Effects of different combinations.}
    In Table \ref{tab:tab3_comb}, we explore the effect of different combinatorial orders of intra- ($\mathcal{A}$: \ding{172} implicit intra-modal fragment interaction and \ding{173} explicit intra-modal fragment interaction) and inter-modal ($\mathcal{B}$: \ding{174} local-local inter-modal interaction and \ding{175} local-global inter-modal interaction) interactions on cross-modal retrieval. Our \textit{Hire} firstly concentrate the relevant information on each target fragment within modality based on the implicit and explicit relationships and then refine the local features based on the cross-level local-local and local-instance attentions, which can improve the semantic representation of each local fragment and further improve later inter-modal interactions with these contextual relationship enhancements.  
    Specifically, when the inter-modal feature interactions are used first and then the intra-modal feature enhancements are used, the retrieval performance drops from 532.6\% to 506.7\% in terms of rSum.
    It suggests that intra-modal interactions integrating potential relationships between the correlated objects into regional features can help the later inter-modal feature interactions obtain more contextual information. 
    Once the order of interactions is reversed, each fragment that obtains contextual information from another modality may be corrupted by subsequent intra-modal interactions, and the original intra-modal relationships will not be accurate based on new contextual object features.
    Furthermore, we change the order of implicit and explicit relationship reasoning module within intra-modal interaction ($\mathcal{A}$: \ding{172}\ding{173}$\rightarrow$\ding{173}\ding{172}) and the order of local-local and local-global cross-modal interactions ($\mathcal{B}$: \ding{174}\ding{175}$\rightarrow$\ding{175}\ding{174}) to evaluate the effectiveness of different combinations of intra-modal interaction and inter-modal interaction, respectively. 
    When the order of implicit and explicit relational reasoning within the modalities is changed, \textit{Hire} decreases its rSum score to 529.8\% on MS-COCO. It suggests that the implicit relational reasoning makes up for the omission of the explicit relationship modelling caused by the scene graph model, thereby improving the fault tolerance of relationship reasoning and model robustness. 
    when changing the order of local-local and local-global inter-modal interactions, the effect of the model does not fluctuate much.

\begin{figure}[t] 
	\centering
 \includegraphics[width=1.0\linewidth]{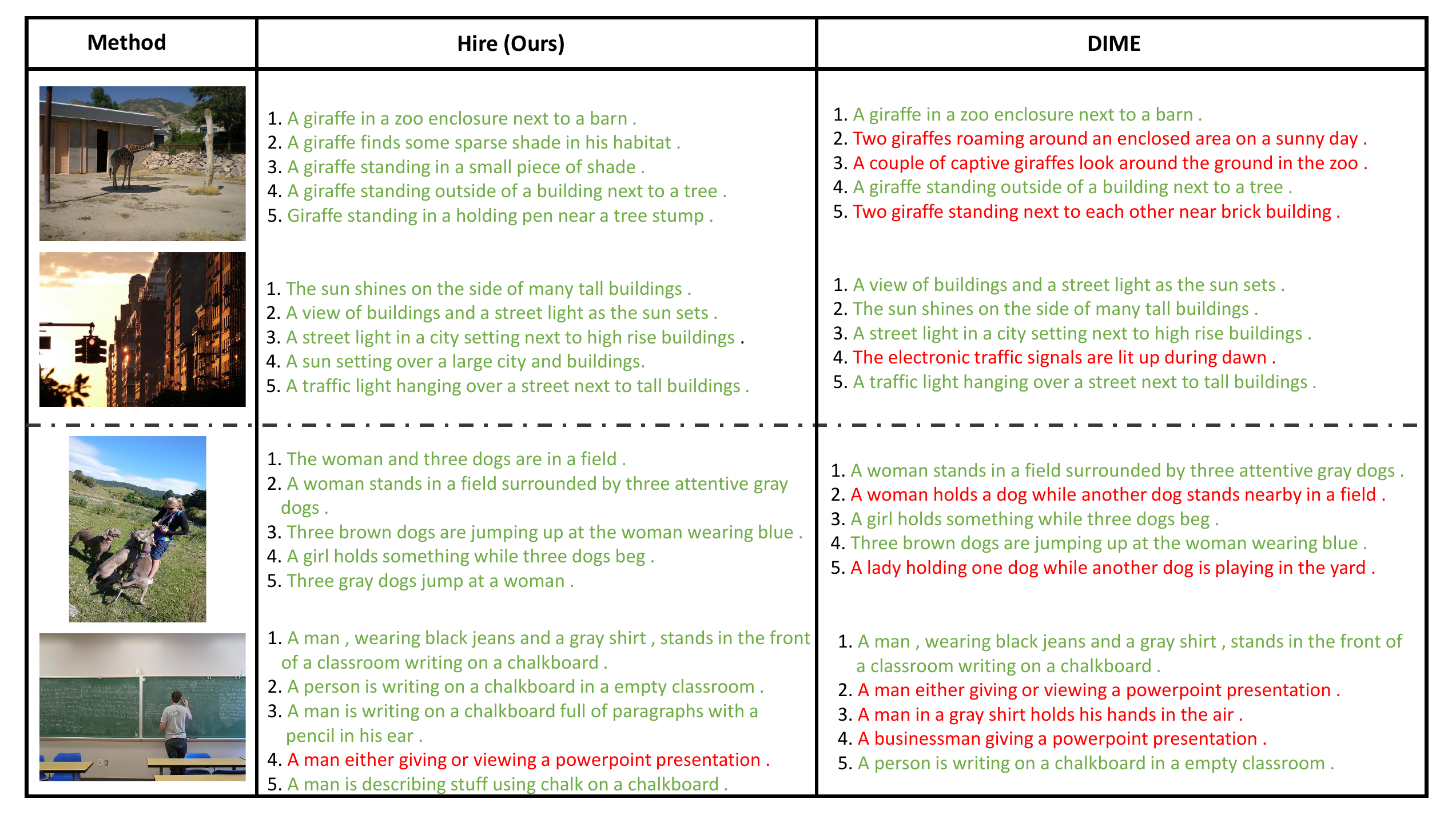}
     \vspace{-2em}
	\caption{Comparisons of image-to-text matching between the proposed \textit{Hire} and DIME [29] on MS-COCO (at the top) and Flickr30K (at the bottom). For each image query,  we present the top-5 retrieved sentences, where the mismatches are highlight in red. }
    	\label{fig:i2t_supply}
    \vspace{-1em}
\end{figure}
\begin{figure}[t] 
	\centering
    \includegraphics[width=1.0\linewidth]{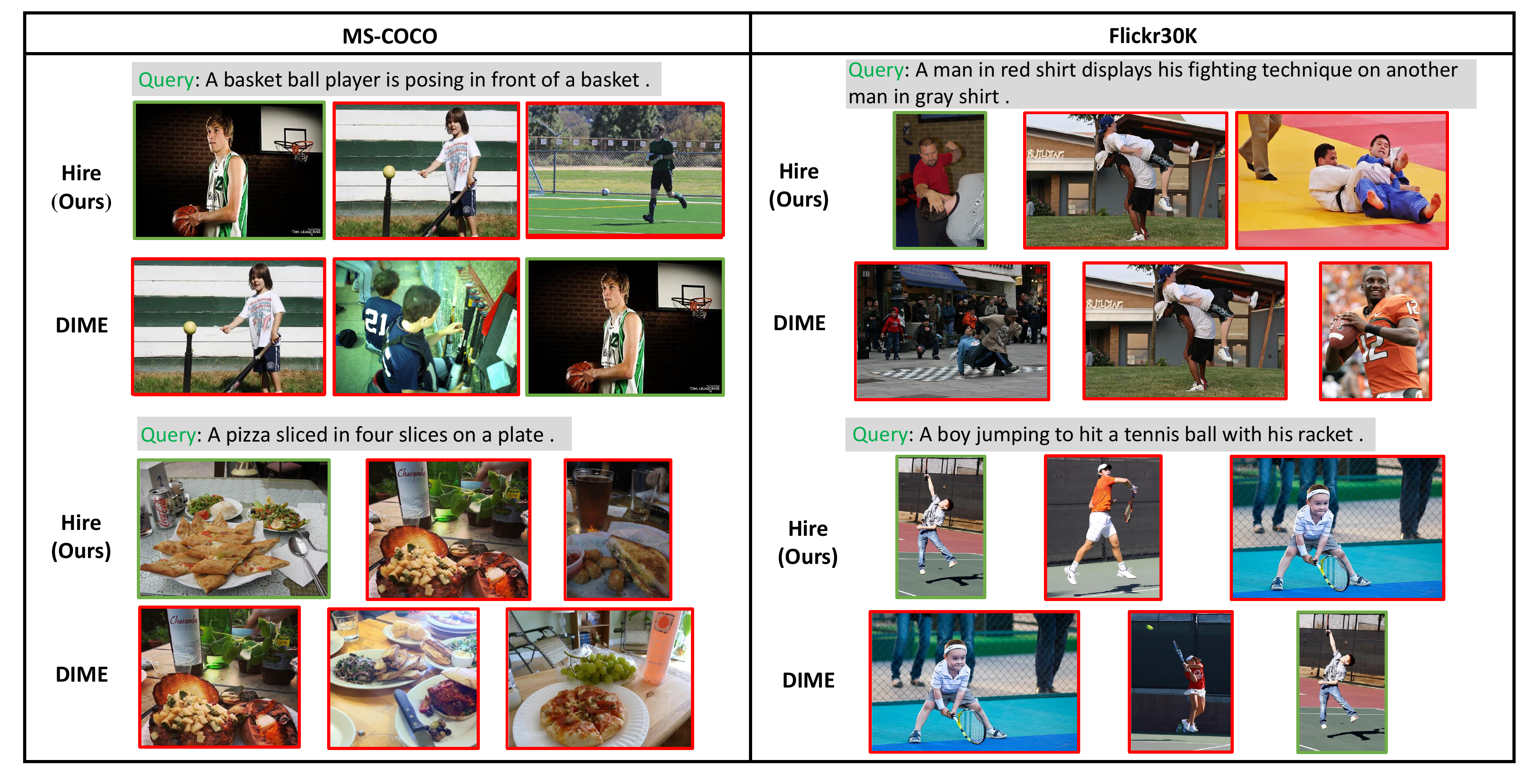}
    \vspace{-2em}
    \caption{Comparisons of text-to-image matching between our \textit{Hire} and DIME [29] on MS-COCO and Flickr30K. For each text query, we present the top 3 ranked images, ranking from left to right. The correctly matched images are marked in green and the mismatched images are marked in red (best viewed in color).}
    	\label{fig:t2i_supply}
     \vspace{-1em}
\end{figure}

    \subsection{Visualization of Results}  
        In Figure \ref{fig:example_Graph_I2TLL_T2ILL_attention}, to better understand the process of intra- and inter-modal interactions of \textit{Hire}, we visualize (i) the refined relationships between each target object and other objects via the implicit and explicit visual object relationship reasoning modules (VSA and VSSG), (ii) the top-4 relevant words corresponding to each object region for image-to-text, and (iii) the top-5 relevant object regions corresponding to each word for text-to-image after local-local inter-modal interaction. 
        As shown in Figure \ref{fig:example_Graph_I2TLL_T2ILL_attention} (i), we have observed that the implicit VSA facilitates the information flow between different regions, but it cannot accurately capture object relationships. 
        The proposed explicit VSSG provides more precise spatial and semantic correlations between the object regions, which can concentrate relevant regional information on the target object in both spatial and semantic levels. 
        The combination of implicit and explicit relationship reasoning contributes to the more comprehensive interaction of cross-modal information in multiple levels. 
        In addition, we also visualize the detailed results of the local-local inter-modal interaction for the relevant pairs on the region-words level (Figure \ref{fig:example_Graph_I2TLL_T2ILL_attention} (ii)) and the word-regions level (Figure \ref{fig:example_Graph_I2TLL_T2ILL_attention} (iii)) guided by VSSG on image-to-text and text-to-image directions, respectively. 
        The results show that the inter-modal interactions accurately calculate the micro fragment correlations of one modality from the other modality, which reflects its ability on effectively narrowing the semantic gap between different modalities.

        To further display the effectiveness of the proposed \textit{Hire}, we show some representative matching results from sentence and image retrieval on both MS-COCO and Flickr30K in Figure \ref{fig:i2t_supply} and Figure \ref{fig:t2i_supply}, respectively. 
        For image-to-text matching, we visualize the top-5 retrieved sentences predicted by our \textit{Hire} and baseline DIME \cite{DIME}, where the mismatches are highlighted in red.
        Furthermore, we show the top-3 ranked images for each sentence in Figure \ref{fig:t2i_supply} by our \textit{Hire} and baseline DIME \cite{DIME}.
        Compared with the state-of-the-art DIME \cite{DIME}, which also utilizes the hybrid-modal interactions, our \textit{Hire} shows stronger retrieval performance in most of hard cases with smaller model parameters. 
        
    \section{Discussion} \label{Discussion}        
          Pre-trained visual language representations on large-scale datasets are becoming increasingly popular, especially in companies with large-scale parallel computing power. However, due to the limitation of computation facility requirements, it is difficult to carry out large-scale pre-training in universities or research institutions.  For example, UNITER-base \citep{li2020unicoder} utilized 882 V100 GPU hours to train a base model and ALIGN \citep{jia2021scaling} used 1024 TPUv3. 
          Our \textit{Hire} uses only one GPU to achieve results that are competitive with mainstream large-scale models. For example, compared with  CLIP\citep{radford2021learning} trained on 400M image-text pairs using over 500 GPUs, our \textit{Hire} can achieve R@1 scores of 61.7\% (+3.3\%) and 45.2\% (+7.4\%) on text retrieval and image retrieval respectively. 
          In addition, this is of great significance for fixed scene matching tasks with small batches of private data, which allows private matching models to be trained without relying on large computing resources.

\section{Conclusion} \label{conclusion}
In this paper, we propose \textit{Hire}, a novel semantic enhanced hybrid-modal interaction method for image-text matching. 
\textit{Hire} engages in (i) enhancing the visual semantic representation with the implicit and explicit inter-object relationships and (ii) enhancing the visual and textual semantic representation with multi-level joint semantic correlations on intra-fragment, inter-fragment, and inter-instance. 
To this end, we propose the hybrid-modal (intra-modal and inter-modal) semantic correlations and advance the integrated structured model with cross-modal semantic alignment in an end-to-end representation learning way.
Extensive quantitative comparisons demonstrate that our \textit{Hire} achieves state-of-the-art performance on most of the standard evaluation metrics across MS-COCO and Flickr30K benchmarks.


\bibliographystyle{ACM-Reference-Format}
\bibliography{sample-base}


\end{document}